# Selecting Uncertainty Calculi and Granularity: An Experiment in Trading-off Precision and Complexity*

Piero P. Bonissone

Keith S. Decker

General Electric Corporate Research and Development
1 River Rd. Bldg. K1, Room 5C32A and 5C38
Schenectady, New York 12301
ARPANET:BONISSONE @GE-CRD
ARPANET: DECKER @GE-CRD

*ABSTRACT*

The management of uncertainty in expert systems has usually been left to *ad hoc* representations and rules of combinations lacking either a sound theory or clear semantics. The objective of this paper is to establish a theoretical basis for defining the syntax and semantics of a small subset of calculi of uncertainty operating on a given term set of linguistic statements of likelihood. Each calculus will be defined by specifying a negation, a conjunction and a disjunction operator. Families of Triangular norms and conorms will provide the most general representations of conjunction and disjunction operators. These families provide us with a formalism for defining an infinite number of different calculi of uncertainty. The term set will define the uncertainty *granularity*, i.e. the finest level of distinction among different quantifications of uncertainty. This granularity will limit the ability to differentiate between two similar operators. Therefore, only a small finite subset of the infinite number of calculi will produce notably different results. This result is illustrated by an experiment where nine different calculi of uncertainty are used with three term sets containing five, nine, and thirteen elements, respectively.

## 1. INTRODUCTION

The aggregation of uncertain information (facts) is a recurrent need in the reasoning process of an expert system. Facts must be aggregated to determine the degree to which the premise of a given rule has been satisfied, to verify the extent to which external constraints have been met, to propagate the amount of uncertainty through the triggering of a given rule, to summarize the findings provided by various rules or knowledge sources or experts, to detect possible inconsistencies among the various sources, and to rank different alternatives or different goals.

In a recent survey of reasoning with uncertainty (Bonissone 1985a, b), it is noted that the presence of uncertainty in reasoning systems is due to a variety of sources: the *reliability* of the information, the inherent *imprecision* of the representation language in which the information is conveyed, the *incompleteness* of the information, and the *aggregation* or summarization of information from multiple sources.

The existing approaches surveyed in that study are divided into two classes: numerical and symbolic representations. The numerical approaches generally tend to impose some restrictions upon the type and structure of the information, e.g. mutual exclusiveness of hypotheses, conditional independence of evidence, etc. These approaches represent uncertainty as a precise quantity (scalar or interval) on a given scale. They require the user or expert to provide a *precise* yet *consistent* numerical assessment of the uncertainty of the atomic data and of their relations. The output produced by these systems is the result of laborious computations, guided by well-defined calculi, and *appears* to be equally precise. However, given the difficulty in consistently eliciting such numerical values from the user, it is clear that these models of uncertainty require an unrealistic level of precision that does not actually represent a real assessment of the uncertainty.

Models based on symbolic representations, on the other hand, are mostly designed to handle the aspect of uncertainty derived from the *incompleteness* of the information. However, they are generally inadequate to handle the case of *imprecise* information, since they lack any measure to quantify confidence levels.

The objective of this paper is to examine the various calculi of uncertainty and to define a rationale for their selection. The number of calculi to be considered will be a function of the uncertainty granularity, i.e. the finest level of distinction among different quantifications of uncertainty that adequately represent the user's discriminating perception. To accomplish this objective we will establish the theoretical framework for defining the syntax of a small subset of calculi of uncertainty operating on a given term set of linguistic statements of likelihood.

In section 2 of this paper, the negation, conjunction, and disjunction operators that form the various calculi of uncertainty are described in terms of their most generic representation: families of functions (Triangular norms and conorms) satisfying the basic axioms expected of set operations such as intersection and union.

In section 3, linguistic variables defined on the [0,1] interval are interpreted as verbal probabilities and their semantics are represented by fuzzy numbers. The term set of linguistic variables defines the granularity of the confidence assessment values that can be consistently expressed by users or experts. A nine element term set is given as an example.

---
* This work was partially supported by the Defense Advanced Research Projects Agency (DARPA) contract F30602-85-C0033. Views and conclusions contained in this paper are those of the authors and should not be interpreted as representing the official opinion or policy of DARPA or the U.S. Government.

Section 4 describes the experiment, consisting in evaluating nine different T-norms with the elements of three different term sets containing five, nine, and thirteen elements, respectively. A review of the techniques required to implement the experiment is also provided. The review covers the implementation of the extension principle, a formalism that enables crisply defined functions to be evaluated with fuzzy-valued arguments, and describes the linguistic approximation, a process required to map the result of the aggregation of two elements of the term set back into the term set.

Section 5 shows the results of computing the closures of selected operators on common term sets. An analysis of the results of this experiment shows the equivalence of some calculi of uncertainty that produce indistinguishable results within the granularity of a given term set. Possible interpretations for the calculi that produce notably different results are suggested in the last part of this section.

Section 6 illustrates the conclusions of this paper.

## 2. AGGREGATION OPERATORS

According to their characteristics, there are three basic classes of aggregation: *conjunctions, trade-offs,* and *disjunctions*. Dubois and Prade (Dubois & Prade. 1984) have shown that Triangular norms (T-norms), averaging operators, and Triangular conorms (T-conorms) are the most general families of binary functions that respectively satisfy the requirements of the conjunction, trade-off, and disjunction operators. T-norms and T-conorms are two-place functions from $[0,1] \times [0,1]$ to $[0,1]$ that are monotonic, commutative and associative. Their corresponding boundary conditions satisfy the truth tables of the logical AND and OR operators. Averaging operators are symmetric and idempotent but are not associative. They do not have a corresponding logical operator since, on the $[0,1]$ interval, they are *located* between the conjunctions and the disjunctions.

The generalizations of conjunctions and disjunctions play a vital role in the management of uncertainty in expert systems: they are used in evaluating the satisfaction of premises, in propagating uncertainty through rule chaining, and in consolidating the same conclusion derived from different rules. More specifically, they provide the answers to the following questions:

— When the premise is composed of multiple clauses, how can we aggregate the degree of certainty $z_i$ of the facts matching the clauses of the premise? i.e. what is the function $T(z_1, \ldots, z_n)$ that determines $z_p$, the degree of certainty of the premise?

— When a rule does not represent a logical implication, but rather an empirical association between premise and conclusion, how can we aggregate the degree of satisfaction of the premise $z_p$ with the strength of the association $s_r$? i.e. what is the function $G(z_p, s_r)$ that propagates the uncertainty through the rule?

— When the same conclusion is established by multiple rules with various degrees of certainty $y_1, \ldots, y_m$, how can we aggregate these contributions into a final degree of certainty? i.e. what is the function $S(y_1, \ldots, y_m)$ that consolidates the certainty of that conclusion?

The following three subsections describe the axiomatic definitions of the conjunction, disjunction, and negation operators.

### 2.1 Conjunction and Propagation Using Triangular Norms

The function $T(a,b)$ aggregates the degree of certainty of two clauses in the same premise. This function performs an INTERSECTION operation and satisfies the conditions of a Triangular Norm (T-norm):

$T(0,0) = 0$ [boundary]
$T(a,1) = T(1,a) = a$ [boundary]
$T(a,b) \leq T(c,d)$ if $a \leq c$ and $b \leq d$ [monotonicity]
$T(a,b) = T(b,a)$ [commutativity]
$T(a, T(b,c)) = T(T(a,b),c)$ [associativity]

The use of a T-norm, a two-place function, to represent the intersection of the clauses in a premise does not limit, however, the number of clauses in such a premise. Because of the associativity of the T-norms, it is possible to define recursively $T(x_1, \ldots, x_n, x_{n+1})$, for $x_1, \ldots, x_{n+1} \in [0,1]$, as:

$$T(x_1, \ldots, x_n, x_{n+1}) = T(T(x_1, \ldots, x_n), x_{n+1})$$

A special case of the conjunction is the *WEIGHTING* function $G(z_p, s_r)$, which attaches a weight or certainty measure to the conclusion of a rule. This weight represents the aggregation of the weight of the premise of the rule $z_p$ (indicating the degree of fulfillment of the premise) with the strength of the rule $s_r$ (indicating the degree of causal implication or empirical association of the rule). This function satisfies the same conditions of the T-norm (although it does not need to be commutative.)

### 2.2 Disjunction Using Triangular Conorms

The function $S(a,b)$ aggregates the degree of certainty of the (same) conclusions derived from two rules. This function performs a UNION operation and satisfies the conditions of a Triangular Conorm (T-conorm):

$$S(1,1) = 1 \qquad \text{[boundary]}$$
$$S(0,a) = S(a,0) = a \qquad \text{[boundary]}$$
$$S(a,b) \leq S(c,d) \text{ if } a \leq c \text{ and } b \leq d \qquad \text{[monotonicity]}$$
$$S(a,b) = S(b,a) \qquad \text{[commutativity]}$$
$$S(a,S(b,c)) = S(S(a,b),c) \qquad \text{[associativity]}$$

A T-conorm can be extended to operate on more than two arguments in a manner similar to the extension for the T-norms. By using a recursive definition, based on the associativity of the T-conorms, we can define:

$$S(y_1, \ldots, y_m, y_{m+1}) = S(S(y_1, \ldots, y_m), y_{m+1})$$

## 2.3 Negation Operators and Calculi of Uncertainty

The selection of a T-norm, Negation operator and T-conorm defines a particular *calculus* of uncertainty. The axioms for a Negation operator have been discussed by several researchers (Bellman & Giertz, 1973; Lowen, 1978; Trillas, 1979). The axioms are:

$$N(0) = 1 \qquad \text{[boundary]}$$
$$N(1) = 0 \qquad \text{[boundary]}$$
$$N(x) > N(y) \text{ if } x < y \qquad \text{[strictly monotonic decreasing]}$$
$$N(a) = \lim_{x \to a} N(x) \qquad \text{[continuity]}$$
$$N(N(x)) = x \qquad \text{[involution]}$$

In most expert systems, a common selection of functions is:

$$CONJUNCTION = T(a,b) = T_3(a,b) = min(a,b) \qquad DISJUNCTION = S(a,b) = S_3(a,b) = max(a,b)$$
$$WEIGHTING = G(a,b) = T_2(a,b) = ab \qquad NEGATION = N(a) = 1-a$$

## 2.4 Relationships between T-norms and T-conorms

For suitable negation operations $N(x)$, such as $N(x)=1-x$, T-norms $T$ and T-conorms $S$ are duals in the sense of the following generalization of DeMorgan's Law:

$$S(a,b) = N(T(N(a), N(b))) \qquad T(a,b) = N(S(N(a), N(b)))$$

This duality implies that the extensions of the intersection and union operators cannot be independently defined and they should therefore analyzed as DeMorgan triples ($T(.,.), S(.,.), N(.)$) or, for a common negation operator like $N(a) = 1-a$, as DeMorgan pairs ($T(.,.), S(.,.)$). Some typical pairs of T-norms $T(a,b)$ and their dual T-conorms $S(a,b)$ are the following:

$$T_0(a,b) = min(a,b) \text{ if } max(a,b)=1 \qquad S_0(a,b) = max(a,b) \text{ if } min(a,b)=1$$
$$\qquad\qquad = 0 \text{ otherwise} \qquad\qquad\qquad\qquad = 1 \text{ otherwise}$$

$$T_1(a,b) = max(0, a+b-1) \qquad S_1(a,b) = min(1, a+b)$$

$$T_{1.5}(a,b) = (ab)/[2-(a+b-ab)] \qquad S_{1.5}(a,b) = (a+b)/(1+ab)$$

$$T_2(a,b) = ab \qquad S_2(a,b) = a+b - ab$$

$$T_{2.5}(a,b) = (ab)/(a+b-ab) \qquad S_{2.5}(a,b) = (a+b-2ab)/(1-ab)$$

$$T_3(a,b) = min(a,b) \qquad S_3(a,b) = max(a,b)$$

These operators are ordered as following:

$$T_0 \leq T_1 \leq T_{1.5} \leq T_2 \leq T_{2.5} \leq T_3 \leq S_3 \leq S_{2.5} \leq S_2 \leq S_{1.5} \leq S_1 \leq S_0$$

Notice that any T-norm $T(a,b)$ and any T-conorm $S(a,b)$ are bounded by:

$$T_0(a,b) \leq T(a,b) \leq T_3(a,b) \qquad S_3(a,b) \leq S(a,b) \leq S_0(a,b)$$

This set of boundaries implies that the averaging operators, used to represent trade-offs are located between the MIN operator $T_3$ (upper bound of T-norms) and the MAX operator $S_3$ (lower bound of T-conorms). These limits have a very intuitive explanation since, if compensations are allowed in the presence of conflicting goals, the resulting trade-off should lie between the most optimistic lower bound and the most pessimistic upper bound, i.e., the worst and best local estimates. Averaging operators are symmetric and idempotent, but, unlike T-norms and T-conorms, are not associative. A detailed description of averaging operators can be found elsewhere (Dubois & Prade, 1984).

## 2.5 Families of T-norms and T-conorms

Sometimes it is desirable to blend some of the previously described T-norm operators in order to smooth some of their effects. While it is always possible to generate a linear combination of two operators, in most cases this would imply giving up the associativity property. However, associativity is the most crucial property of the T-norms (Schweizer & Sklar, 1963; Ling 1965) since it allows the decomposition of multiple-place functions in terms of two-place functions. The correct solution is to find a family of T-norms that ranges over the desired operators. The proper selection of a parameter will then define the intermediate operator with the desired effect while still preserving associativity.

There are at least six families of T-norms $T_s(a,b,p)$ with their dual* T-conorms $S_s(a,b,p)$. The value of the subscript $s$ will denote the family of norms; $p$, the third argument of each norm, will denote the parameter used by the corresponding family.

YAGER: $T_Y(a,b,q) = 1 - \text{MIN} \{1, [(1-a)^q + (1-b)^q]^{1/q}\}$     for $q > 0$
YAGER: $S_Y(a,b,q) = \text{MIN} \{1, (a^q + b^q)^{1/q}\}$     for $q > 0$

DUBOIS: $T_D(a,b,\alpha) = (ab)/\text{MAX} \{a,b,\alpha\}$     for $\alpha \in [0,1]$
DUBOIS: $S_D(a,b,\alpha) = [a+b-ab - \text{MIN} \{a,b,(1-\alpha)\}]/\text{MAX} \{(1-a),(1-b),\alpha\}$     for $\alpha \in [0,1]$

HAMACHER: $T_H(a,b,\gamma) = (ab)/[\gamma+(1-\gamma)(a+b-ab)]$     for $\gamma \geq 0$
HAMACHER: $S_H(a,b,\gamma) = [a+b+(\gamma-2)ab]/[1+(\gamma-1)ab]$     for $\gamma \geq 0$

SCHWEIZER: $T_{Sc}(a,b,p) = \text{MAX} \{0, (a^{-p}+b^{-p}-1)\}^{-1/p}$     for $p \in [-\infty,\infty]$
SCHWEIZER: $S_{Sc}(a,b,p) = 1 - \text{MAX} \{0, [(1-a)^{-p}+(1-b)^{-p}-1]\}^{-1/p}$     for $p \in [-\infty,\infty]$

FRANK: $T_F(a,b,s) = \text{Log}_s [1+(s^a-1)(s^b-1)/(s-1)]$     for $s > 0$
FRANK: $S_F(a,b,s) = 1 - \text{Log}_s [1+(s^{(1-a)}-1)(s^{(1-b)}-1)/(s-1)]$     for $s > 0$

SUGENO: $T_{Su}(a,b,\lambda) = \text{MAX} \{0, (\lambda+1)(a+b-1) -\lambda ab\}$     for $\lambda \geq -1$
SUGENO: $S_{Su}(a,b,\lambda) = \text{MIN} \{1, a+b-\lambda.a.b\}$     for $\lambda \geq -1$

The above families of T-norms and T-conorms are individually described in the literature (Yager, 1980; Dubois & Prade, 1982; Hamacher, 1975; Schweizer & Sklar, 1963; Frank, 1979; Sugeno, 1974, 1977). The following table indicates the value of the parameter for which the above families of norms reproduce the most common T-norms $\{T_0, \ldots, T_3\}$.

| $T_Y(a,b,q)$ | $T_D(a,b,\alpha)$ | $T_H(a,b,\gamma)$ | $T_{Sc}(a,b,p)$ | $T_F(a,b,s)$ | $T_{Su}(a,b,\lambda)$ | T-norm |
|---|---|---|---|---|---|---|
| $q$ | $\alpha$ | $\gamma$ | $p$ | $s$ | $\lambda$ | |
| $\to 0^+$ | . | $\to \infty$ | $\to -\infty$ | . | $\to \infty$ | $T_0$ |
| 1 | . | -1 | . | $\to \infty$ | 0 | $T_1$ |
| . | . | 2 | . | . | . | $T_{1.5}$ |
| . | 1 | 1 | $\to 0$ | $\to 1$ | -1 | $T_2$ |
| . | . | 0 | . | . | . | $T_{2.5}$ |
| $\to \infty$ | 0 | . | $\to \infty$ | $\to 0^+$ | . | $T_3$ |

TABLE 1: *Ranges of the six parametrized families of T-norms*

The vertical bars "|" used in Table 1 indicate the legal ranges of each parameter. The table for the T-conorms is identical to the above except for the header, where the families of T-norms are replaced by the corresponding families of T-conorms, and the last column, where the T-norms are replaced by their respective dual T-conorms, i.e. $T_0$ by $S_0$, etc.

## 3. LINGUISTIC VARIABLES DEFINED ON THE INTERVAL [0,1]

These families of norms can specify an infinite number of calculi that operate on arguments taking *real number* values on the [0,1] interval. This *fine-tuning* capability would be useful if we needed to compute, with a high degree of precision, the results of aggregating information characterized by very precise measures of its uncertainty. However, when users or experts must provide these measures, an assumption of *fake precision* must usually be made to satisfy the requirements of the selected calculus.

Szolovits and Pauker (Szlovits & Pauker, 1978) noted that "...while people seem quite prepared to give qualitative estimates of likelihood, they are often notoriously unwilling to give precise numerical estimates to outcomes." This seems to indicate that any scheme that relies on the user providing *consistent* and *precise numerical* quantifications of the confidence level of his/her conditional or unconditional statements is bound to fail.

---
* The dual T-conorms are obtained from the T-norm by using the generalized DeMorgan's Law with negation defined by $N(x)=1-x$. This negation operator, however, is not unique as illustrated by Lowen (Lowen, 1978).

It is instead reasonable to expect the user to provide *linguistic* estimates of the likelihood of given statements. The experts and users would be presented with a verbal scale of certainty expressions that they could then use to describe their degree of certainty in a given rule or piece of evidence. Recent psychological studies have shown the feasibility of such an approach: "...A verbal scale of probability expressions is a compromise between people's resistance to the use of numbers and the necessity to have a common numerical scale" (Beyth-Marom, 1982).

Each linguistic likelihood assessment is internally represented by fuzzy intervals, i.e., fuzzy numbers. A fuzzy number is a fuzzy set (Zadeh, 1965) defined on the real line. In this case, the membership function of a fuzzy set defined on a truth space, i.e. the interval $[0,1]$, could be interpreted as the *meaning* of a label describing the degree of certainty in a linguistic manner (Zadeh, 1975; Bonissone, 1980). During the aggregation process, these fuzzy numbers will be modified according to given combination rules and will generate another membership distribution that could be mapped back into a linguistic term for the user's convenience or to maintain closure. This process, referred to as linguistic approximation, has been extensively studied (Bonissone, 1979a, b) and will be briefly reviewed in section 4.2.

### 3.1 Example of a Term Set of Linguistic Probabilities

Let us consider the following term set $L_2$:

{*impossible extremely_unlikely very_low_chance small_chance it_may meaningful_chance most_likely extremely_likely certain*}

Each element $E_i$ in the above term set represents a statement of linguistic probability or likelihood. The semantics of each element $E_i$ are provided by a fuzzy number $N_i$ defined on the $[0,1]$ interval. A fuzzy number $N_i$ can be described by its continuous membership function $\mu_{N_i}(x)$, for $x \in [0,1]$.

A computationally more efficient way to characterize a fuzzy number is to use a parametric representation of its membership function. This parametric representation (Bonissone, 1980) is achieved by the 4-tuple $(a_i, b_i, \alpha_i, \beta_i)$. The first two parameters indicate the interval in which the membership value is 1.0; the third and fourth parameters indicate the left and right *width* of the distribution. Linear functions are used to define the slopes. Therefore, the membership function $\mu_{N_i}(x)$, of the fuzzy number $N_i = (a_i, b_i, \alpha_i, \beta_i)$ is defined as follows:

$$\begin{aligned}
\mu_{N_i}(x) &= 0 & \text{for } x < (a_i - \alpha_i) \\
&= (1/\alpha_i)(x - a_i + \alpha_i) & \text{for } x \in [(a_i - \alpha_i), a_i] \\
&= 1 & \text{for } x \in [a_i, b_i] \\
&= (1/\beta_i)(b_i + \beta_i - x) & \text{for } x \in [b_i, (b_i + \beta_i)] \\
&= 0 & \text{for } x > (b_i + \beta_i)
\end{aligned}$$

Figure 1 shows the membership distribution of the fuzzy number $N_i = (a_i, b_i, \alpha_i, \beta_i)$.

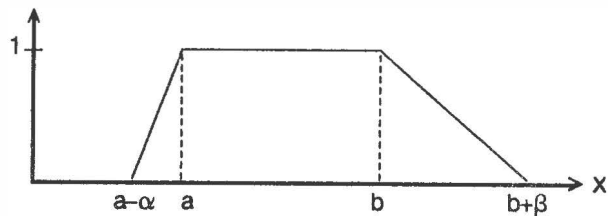

**FIGURE 1**: *Membership Distributions of $N_i = (a_i, b_i, \alpha_i, \beta_i)$*

The second column of Table 2 indicates the semantics of the proposed term set $L_2$. The membership distributions of the term set elements are illustrated in Figure 2. The values of the fuzzy interval associated with each element in the proposed term set were derived from an adaptation of the results of psychological experiments on the use of linguistic probabilities (Beyth-Marom, 1982). For most of the elements in the term set, the two measures of dispersions used by Beyth-Marom, e.g., the interquartile range ($C_{25}$-$C_{75}$) and the 80 per cent range ($C_{10}$-$C_{90}$), were used to define respectively the intervals $[a_i, b_i]$ and $[(a_i - \alpha_i), (b_i + \beta_i)]$ of each fuzzy number $N_i$.

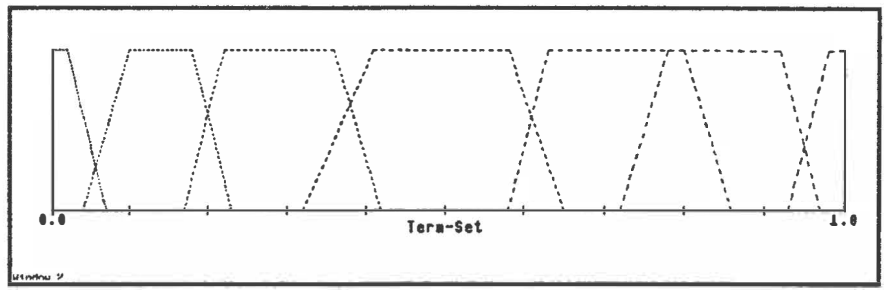

**FIGURE 2**: *Membership Distributions of Elements in $L_2$*

# 4. DESCRIPTION OF THE EXPERIMENT AND REQUIRED TECHNIQUES

## 4.1 The Experiment

The experiment consists in selecting nine different T-norms that, in combination with their DeMorgan dual T-conorms and a negation operator, define nine different calculi of uncertainty. Three different term sets--containing five, nine, and thirteen elements--provide three different levels of granularity for quantifying the uncertainty. The T-norms will be evaluated with all the elements of the three term sets; their results will be compared and analyzed for significant differences.

To select the T-norms for the experiment, we first took the three most important T-norms, i.e., $T_1$, $T_2$, $T_3$, which provide the lower bound of the copulas, an intermediate value, and the upper bound of the T-norms. We then used a parametrized family of T-norms capable of covering the entire spectrum between $T_1$ and $T_3$. Our choice fell on the family of T-norms proposed by Schweizer and Sklar, i.e., $T_{Sc}(a,b,p)$, described in section 2.4. The selection of this particular family of T-norms was due to its full coverage of the spectrum and its numerical stability in the neighborhood of the origin. We then selected six values of the parameter $p$ to probe the space between $T_1$ and $T_2$ ($p \in [-1,0]$), and between $T_2$ and $T_3$ ($p \in [0,\infty]$). The six T-norms instantiated from this family were: $T_{Sc}(a,b,-0.8)$, $T_{Sc}(a,b,-0.5)$, $T_{Sc}(a,b,-0.3)$, $T_{Sc}(a,b,0.5)$, $T_{Sc}(a,b,1)$, $T_{Sc}(a,b,2)$.

The term sets used to provide the different levels of granularity are: $L_1$, $L_2$, $L_3$.

$L_1$ and $L_3$ contain five and thirteen elements, respectively. Their labels and semantics are defined in the first and third columns in Table 2.

| impossible | (0 0 0 0) | impossible | (0 0 0 0) | impossible | (0 0 0 0) |
|---|---|---|---|---|---|
| unlikely | (0 .25 0 .1) | extremely_unlikely | (0 .02 0 .05) | extremely_unlikely | (0 .02 0 .05) |
| maybe | (.4 .6 .1 .1) | very_low_chance | (.1 .18 .06 .05) | not_likely | (.05 .15 .03 .03) |
| likely | (.75 1 .1 0) | small_chance | (.22 .36 .05 .06) | very_low_chance | (.1 .18 .06 .05) |
| certain | (1 1 0 0) | it_may | (.41 .58 .09 .07) | small_chance | (.22 .36 .05 .06) |
| | | meaningful_chance | (.63 .80 .05 .06) | it_may | (.41 .58 .09 .07) |
| | | most_likely | (.78 .92 .06 .05) | likely | (.53 .69 .09 .12) |
| | | extremely_likely | (.98 1 .05 0) | meaningful_chance | (.63 .80 .05 .06) |
| | | certain | (1 1 0 0) | high_chance | (.75 .87 .04 .04) |
| | | | | most_likely | (.78 .92 .06 .05) |
| | | | | very_high_chance | (.87 .96 .04 .03) |
| | | | | extremely_likely | (.98 1 .05 0) |
| | | | | certain | (1 1 0 0) |
| Term Set $L_1$ | | Term Set $L_2$ | | Term Set $L_3$ | |

**TABLE 2**: *Elements of Terms Sets $L_1$, $L_2$ and $L_3$*

The above experiment can be performed only if some particular computational techniques are used. It is necessary to evaluate the selected T-norms (crisply defined functions) with the elements of the term sets (linguistic variables with fuzzy-valued semantics). Furthermore, the result of this evaluation must be another element of the term set. This implies that closure must be maintained under the application of each T-norm. The following two subsections describe the techniques necessary to satisfy these requirements.

## 4.2 The Extension Principle

The extension principle (Zadeh, 1975) allows any non-fuzzy function to be fuzzified in the sense that if the function arguments are made fuzzy sets, then the function value is also a fuzzy set whose membership function is uniquely specified.

The extension principle states that if the scalar function, $f$, takes $n$ arguments $(x_1, x_2, \ldots, x_n)$, denoted by $X$ and if the membership functions of these arguments are denoted by $\mu_1(x_1), \mu_2(x_2), \ldots, \mu_1(x_n)$, then:

$$\mu_{f(X)}(y) = \sup_{\substack{X \\ \text{s.t. } f(X) = y}} [\inf_{i=1}^{n} \mu_i(x_i)]$$

where SUP and INF denote the *Supremum* and *Infimum* operators.

The use of this formal definition entails various types of computational difficulties (Bonissone, 1980). The solution to these difficulties is based on the parametric representation of the membership distribution of a fuzzy number, i.e $N_i = (a_i, b_i, \alpha_i, \beta_i)$, described in section 3.1. Such a representation allows one to describe uniformly a *crisp number*, e.g., $(a_i, a_i, 0, 0)$; a *crisp interval*, e.g., $(a_i, b_i, 0, 0)$; a *fuzzy number*, e.g., $(a_i, a_i, \alpha_i, \beta_i)$; and a *fuzzy interval* $(a_i, b_i, \alpha_i, \beta_i)$.

The adopted solution consists of deriving the *closed-form* parametric representation of the result. This solution is a very good approximation of the result obtained from using the extension principle to evaluate arithmetic functions with fuzzy numbers, and has a much more limited computational overhead. The formulae providing the closed form solution for inverse, logarithm, addition, subtraction, multiplication, division, and power can be found in the extended version of this paper (Bonissone & Decker, 1985c). These formulae were used in the implementation of the experiment described in section 4.1.

### 4.3 Linguistic Approximation

The process of *linguistic approximation* consists of finding a *label* whose meaning is the same or the closest (according to some metric) to the meaning of an unlabelled membership function generated by some computational model. Bonissone (Bonissone, 1979a, b) has discussed the general solution to this problem.

For the particular case of our experiment, this process was simplified by the small cardinality of the term sets. Therefore, a simplified solution was adopted. From each element of the term set and from the unlabelled membership function representing the result of some arithmetic operation, two features were extracted: the first moment of the distribution and the area under the curve. A weighted Euclidean distance, where the weights reflected the relevance of the two parameters in determining semantic similarity, provided the metric required to select the element of the term set that more closely represented the result.

This process was used in the experiment described in Section 4.1 to provide *closure* under the application of the various T-norms. The closure requirement is required by any calculus of uncertainty to maintain the form and meaning of the linguistic confidence measures throughout the rule chaining and aggregation process.

## 5. EXPERIMENT RESULTS AND ANALYSIS

### 5.1 Tabulated Results

Selected results of the experiment are shown in tabular form in Tables 3, 4, and 5. Each table illustrates the effects of applying $T_1$, $T_2$, and $T_3$ to the elements of a particular term set. Because of the commutativity property of the T-norms, the tables are symmetric.

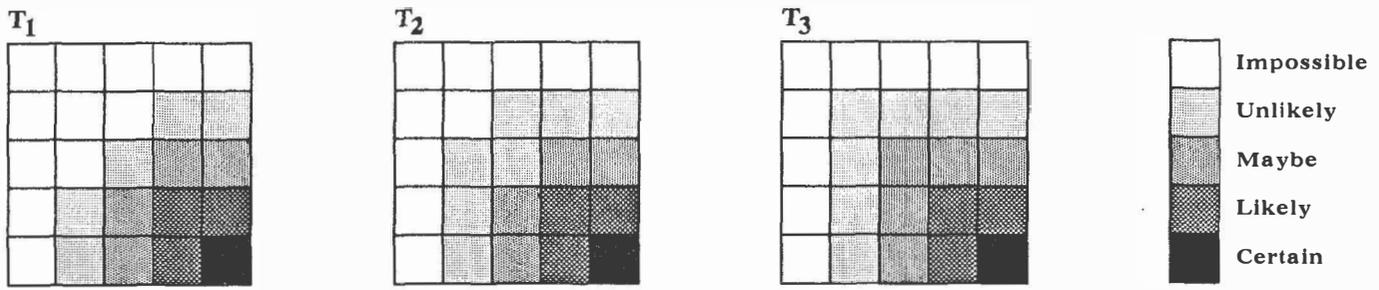

TABLE 3: *Closure of $T_1$, $T_2$, $T_3$ on $L_1$*

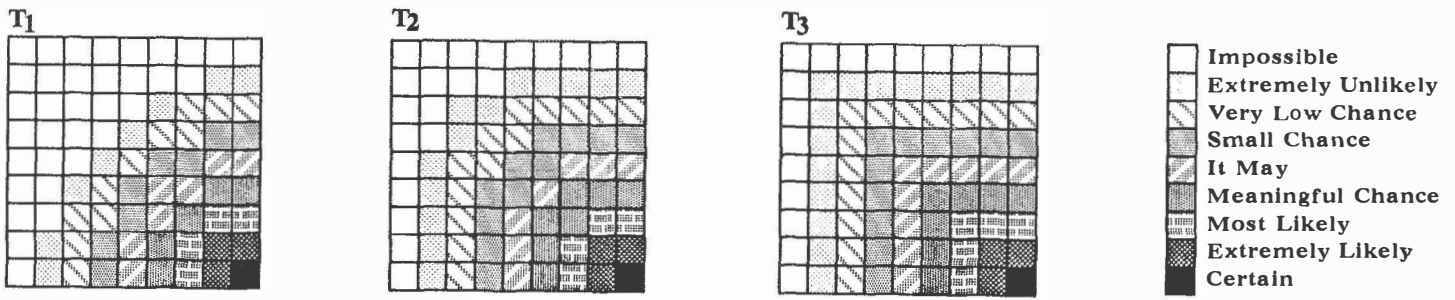

TABLE 4: *Closure of $T_1$, $T_2$, $T_3$ on $L_2$*

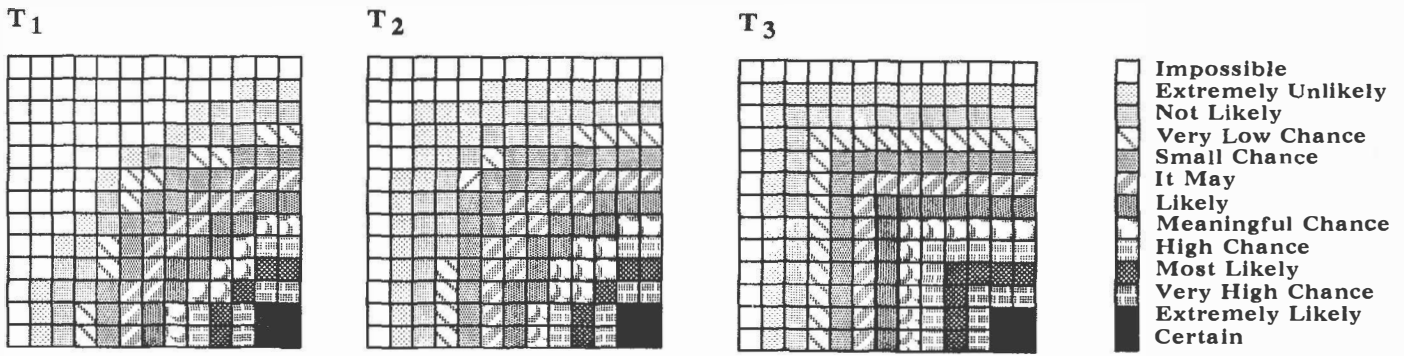

TABLE 5: *Closure of $T_1$, $T_2$, $T_3$ on $L_3$*

## 5.2 Analysis of the Results of the Experiment

The three previous tables graphically illustrate the different behaviors of $T_1$, $T_2$ and $T_3$ when applied to a common term set. As expected $T_1$ was the strictest operator and $T_3$ was the most liberal operator. However, the interesting aspect of the experiment was not rediscovering the behavior of the two extremes but determining how many different variations of behavior we had to consider from the operators located between $T_1$ and $T_3$.

The closures of seven T-norms, bounded by $T_1$ from below and by $T_3$ from above, were computed and compared with the closures of the two extremes. For each of the three term sets, each element in the closure of a given T-norm, i.e., $T_i(E_i, E_j)$, was compared with the same element in the closure of a different T-norm, i.e., $T_k(E_i, E_j)$. The number of differences found by moving from one T-norm to the next was tabulated for each term set and the results shown in Tables 6, 7, and 8. The percentages contained by these tables were computed as the ratio of the number of changes divided by the cardinality of the closure for each term set. Since the closures were symmetric due to the commutativity property of the T-norms, the cardinality of the closure for a term set with $n$ elements was considered to be $n(n+1)/2$.

By analyzing Table 6, it is evident that *no differences* were found among some of the intermediate T-norms. There are indeed three equivalence classes of T-norms producing different results when applied to elements of $L_1$. These classes of equivalence are:

$\{T_1(a,b), T_{Sc}(a,b,-0.8), T_{Sc}(a,b,-0.5), T_{Sc}(a,b,-0.3)\}$, $\{T_2(a,b), T_{Sc}(a,b,0.5)\}$, $\{T_{Sc}(a,b,1), T_{Sc}(a,b,2), T_3(a,b)\}$

From Table 7, we can observe that *few significant differences* were found among the intermediate T-norms when applied to elements of $L_2$. To create equivalence classes among the T-norms, we need to establish a threshold value indicating the maximum percentage of differences that we are willing to tolerate among T-norms of the same class of equivalence. With a threshold of 6.5% we find five classes:

$\{T_1(a,b), T_{Sc}(a,b,-0.8)\}$, $\{T_{Sc}(a,b,-0.5)\}$, $\{T_{Sc}(a,b,-0.3), T_2(a,b)\}$, $\{T_{Sc}(a,b,0.5), T_{Sc}(a,b,1), T_{Sc}(a,b,2)\}$, $\{T_3(a,b)\}$

With a threshold of 15.5% we find three classes:

$\{T_1(a,b), T_{Sc}(a,b,-0.8), T_{Sc}(a,b,-0.5)\}$, $\{T_{Sc}(a,b,-0.3), T_2(a,b)\}$, $\{T_{Sc}(a,b,0.5), T_{Sc}(a,b,1), T_{Sc}(a,b,2), T_3(a,b)\}$

Finally, from Table 8 we can observe that a *larger number of differences* were found among the intermediate T-norms. Using a threshold of 12% we find five classes of equivalence:

$\{T_1(a,b), T_{Sc}(a,b,-0.8)\}$, $\{T_{Sc}(a,b,-0.5)\}$, $\{T_{Sc}(a,b,-0.3), T_2(a,b)\}$, $\{T_{Sc}(a,b,0.5), T_{Sc}(a,b,1), T_{Sc}(a,b,2)\}$, $\{T_3(a,b)\}$

In summary, we can see that three T-norms are sufficient to define the relevant calculi using the five element term set $L_1$; five T-norms are required to represent (88% of the time) the variations in relevant calculi for the thirteen element term set $L_9$. For the case of $L_2$, the same three T-norms used for $L_1$ will suffice if we are willing to accept results that might be slightly different 15.5% of the time. Otherwise, we will have to use five T-norms, as for $L_9$, to reduce the number of slight differences to 6.5%.

For any practical purpose, the three classes of equivalence represented by $T_1$, $T_2$ and $T_3$ more than adequately represent the variations of calculi that can produce different results when applied to elements of term sets with at most nine elements.

The appropriate selection of uncertainty granularity, i.e., the term set cardinality, is still a matter of subjective judgement. However, if we use the very well-known results on *the span of absolute judgement* (Miller, 1967), it seems unlikely that any expert or user could consistently quantify uncertainty using more than nine different values.

| $T_1(a,b)$ | $T_{Sc}(a,b,-.8)$ | $T_{Sc}(a,b,-.5)$ | $T_{Sc}(a,b,-.3)$ | $T_2(a,b)$ | $T_{Sc}(a,b,.5)$ | $T_{Sc}(a,b,1)$ | $T_{Sc}(a,b,2)$ | $T_3(a,b)$ |
|---|---|---|---|---|---|---|---|---|
| 0 / 0% | 0 / 0% | 0 / 0% | 1 / 6.5% | 0 / 0% | 2 / 13.3% | 0 / 0% | 0 / 0% | |
| 0 / 0% | | | | 0 / 0% | | | 0 / 0% | |

**TABLE 6:** *Number of Differences Among The Nine T-norms Using $L_1$*

| $T_1(a,b)$ | $T_{Sc}(a,b,-.8)$ | $T_{Sc}(a,b,-.5)$ | $T_{Sc}(a,b,-.3)$ | $T_2(a,b)$ | $T_{Sc}(a,b,.5)$ | $T_{Sc}(a,b,1)$ | $T_{Sc}(a,b,2)$ | $T_3(a,b)$ |
|---|---|---|---|---|---|---|---|---|
| 1 / 2.2% | 6 / 13.3% | 6 / 13.3% | 3 / 6.5% | 4 / 8.8% | 2 / 4.4% | 1 / 2.2% | 4 / 8.8% | |
| 1 / 2.2% | | | 3 / 6.5% | | | 3 / 6.5% | | |
| 7 / 15.5% | | | 3 / 6.5% | | | 7 / 15.5% | | |

**TABLE 7:** *Number of Differences Among The Nine T-norms Using $L_2$*

| $T_1(a,b)$ | $T_{Sc}(a,b,-.8)$ | $T_{Sc}(a,b,-.5)$ | $T_{Sc}(a,b,-.3)$ | $T_2(a,b)$ | $T_{Sc}(a,b,.5)$ | $T_{Sc}(a,b,1)$ | $T_{Sc}(a,b,2)$ | $T_3(a,b)$ |
|---|---|---|---|---|---|---|---|---|
| 8 | 14 | 15 | 5 | 14 | 7 | 4 | 14 | |
| 8.8% | 15.3% | 16.4% | 5.5% | 15.3% | 7.6% | 4.3% | 15.3% | |
| 8 | | | 5 | | 11 | | | |
| 8.8% | | | 5.5% | | 12% | | | |

**TABLE 8:** *Number of Differences Among The Nine T-norms Using $L_9$*

### 5.3 Meaning of $T_1$, $T_2$, $T_3$

$T_1$, $T_2$ and $T_3$ were the three operators that produced notably different results for $L_4$ and $L_9$. A challenging task is to establish the meaning of each T-norm, i.e., the rationale for selecting one T-norm over the other two.

A first interpretation indicates that $T_1$ seems appropriate to perform the intersection of lower probability bounds (Dempster, 1967) or degrees of *necessity* (Zadeh, 1979). The strict behavior of $T_1$ is required when taking the intersection of *hard evidence*. This choice is in perfect agreement with the fact that the union of lower bounds is superadditive (Zadeh 1979), as is $S_1$, the T-conorm dual of $T_1$. Similarly, $T_3$ is appropriate to represent the intersection of upper probability bounds, or degrees of *possibility*. $T_2$ is the classical probabilistic operator that assumes *independence* of the arguments; its dual T-conorms, $S_2$ is the usual *additive* measure for the union.

To provide a better understanding of these T-norms, we will paraphrase an example introduced by Zadeh (Zadeh, 1983):

> *If 30% of the students in a college are engineers, and 80% of the students are male, how many students are both male and engineers? Although we started with numerical quantifiers, the answer is no longer a number, but is given by the interval [10%, 30%]*

The lower bound of the answer is provided by $T_1(0.3, 0.8)$; $T_3(0.3, 0.8)$ generates its upper bound. $T_2(0.3, 0.8)$ gives a somewhat arbitrary estimate of the answer, based on the independence of the two pieces of evidence.

In Figure 3, we try to describe geometrically the meaning of the three T-norms. The figure illustrates the result of $T_1(0.3, 0.8)$, $T_2(0.3, 0.8)$, and $T_3(0.3, 0.8)$. $T_1$ captures the notion of *worst case*, where the two arguments are considered as *mutually exclusive* as possible (the dimensions on which they are measured are 180° apart). $T_2$ captures the notion of *independence* of the arguments (their dimensions are 90° apart). $T_3$ captures the notion of *best case*, where one of the arguments attempts to *subsume* the other one (their dimensions are collinear, i.e., 0° apart).

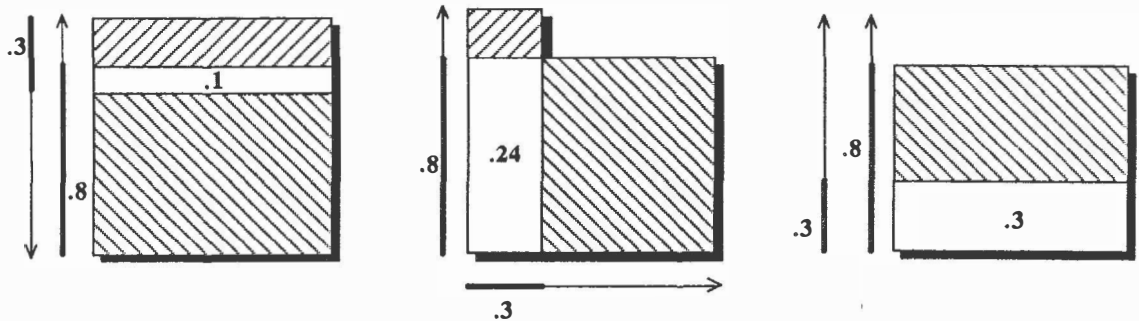

**FIGURE 3:** *Geometrical Interpretation of $T_1(0.3, 0.8)$, $T_2(0.3, 0.8)$, and $T_3(0.3, 0.8)$*

### 6. CONCLUSIONS

In this paper we have presented a formalism to represent any calculus of uncertainty in terms of a selection of a negation operator and two elements from families of T-norms and T-conorms. Because of our skepticism regarding the realism of the *fake precision* assumption required by most existing numerical approaches, we proposed the use of a term set that determines the finest level of specificity, i.e., the *granularity*, of the measure of certainty that the user/expert can *consistently* provide. The suggested semantics for the elements of the term set are given by fuzzy numbers on the [0,1] interval. The values of the fuzzy numbers were determined on the basis of the results of a psychological experiment aimed at the consistent use of linguistic probabilities.

We then proceeded to perform an experiment to test the required level of discrimination among the various calculi, given a fixed uncertainty granularity. We reviewed the techniques required to implement the experiment, such as the extension principle (that permits the evaluation of crisply defined function with fuzzy arguments), a parametric representation of fuzzy numbers (that allows closed form solutions for arithmetical operations), and the process of *linguistic approximation* of a fuzzy number (that guarantees *closure* of the term set under the various calculi of uncertainty).

We computed the closure of nine T-norm operators applied to three different term sets. We analyzed the sensitivity of each operator with respect to the granularity of the elements in the term set; and we finally determined that only three T-norms-- $T_1$, $T_2$, and $T_3$-generated sufficiently distinct results for those term sets that do not contain more than nine elements.